\documentclass{article}

\usepackage{arxiv}

\usepackage[utf8]{inputenc} 
\usepackage[T1]{fontenc}    
\usepackage{hyperref}       
\usepackage{url}            
\usepackage{booktabs}       
\usepackage{amsfonts}       
\usepackage{nicefrac}       
\usepackage{microtype}      
\usepackage{lipsum}
\usepackage{graphicx}
\graphicspath{ {./images/} }
\usepackage{colortbl}
\usepackage{listings}

\title{esCorpius: A Massive Spanish Crawling Corpus}

\author{
Asier Gutiérrez-Fandiño \\
  LHF Labs\\
  \texttt{asier@lhf.ai} \\
   \And
 David Pérez-Fernández\\
  Universidad Autónoma de Madrid\\
  \texttt{david.perez@inv.uam.es} \\
  \And
  Jordi Armengol-Estapé\\
  LHF Labs\\
  University of Edinburgh\\
  \texttt{jordi@lhf.ai}\\
  \And
 David Griol\\
  LHF Labs\\ Universidad de Granada\\
  \texttt{dgriol@ugr.es} \\
  \And
 Zoraida Callejas\\
  LHF Labs\\ Universidad de Granada\\
  \texttt{zoraida@ugr.es} \\
}

\begin{document}
\maketitle

\begin{abstract}
In the recent years, transformer-based models have lead to significant advances in language modelling for natural language processing. However, they require a vast amount of data to be (pre-)trained and there is a lack of corpora in languages other than English. Recently, several initiatives have presented multilingual datasets obtained from automatic web crawling. However, the results in Spanish present important shortcomings, as they are either too small in comparison with other languages, or present a low quality derived from sub-optimal cleaning and deduplication. In this paper, we introduce \textsc{esCorpius}, a Spanish crawling corpus obtained from near 1 Pb of Common Crawl data. It is the most extensive corpus in Spanish with this level of quality in the extraction, purification and deduplication of web textual content. Our data curation process involves a novel highly parallel cleaning pipeline and encompasses a series of deduplication mechanisms that together ensure the integrity of both document and paragraph boundaries. Additionally, we maintain both the source web page URL and the WARC shard origin URL in order to complain with EU regulations. \textsc{esCorpius} has been released under CC BY-NC-ND 4.0 license and is available on HuggingFace.
\end{abstract}

\section{Introduction}
Deep Learning (DL) models in Natural Language Processing (NLP) have achieved unseen performance and thus largely replaced conventional machine learning approaches across multiple applications, such as machine translation, natural language understanding, or natural language generation. One of the main tasks in these areas is language modelling, for which the current state of the art is to use models pre-trained on a data-rich tasks or languages that are subsequently fine-tuned to the target language or task at hand, so that stakeholders do not have to perform expensive pre-training themselves.

Despite the positive results reported in the literature for English pre-trained models, many voices highlight the disparity in the availability and quality of models and data for other languages. As pointed out in \cite{otter_survey_2021}, most NLP research is conducted in English, followed by Mandarin Chinese, and at a great distance to other languages, including Spanish despite its large number of speakers around the world. This situation has negative repercussions on the development of fair NLP technology for all \cite{mehrabininareh_survey_2021}, e.g. disparate access to clinical NLP for speakers of different languages \cite{wu_deep_2020}. 

To address this issue, language models are being trained on monolingual corpora in different languages. For example Multilingual BERT\footnote{\url{https://github.com/google-research/bert/blob/master/multilingual.md}}, which has been recently outperformed by mT5 \cite{xue_mt5_2021}, a massively multilingual model trained with mC4, a dataset of text in 101 languages. However, the quality of mC4 is not as good for non-English languages\footnote{According to one of the authors: \url{https://github.com/allenai/allennlp/discussions/5265\#discussioncomment-2596110}}, which is making some researchers produce clean excerpts of mC4 corresponding to their language of interest, see e.g. the clean Italian mC4 \cite{sarti-nissim-2022-it5}. As warned in \cite{kreutzer_quality_2022}, low-quality data can have pernicious effects, not only in the quality of the results produced, but also because it may lead to the false idea that languages different from English, in our case Spanish, are well enough represented with sufficient high-quality resources.

There also exist parallel corpora, which exploit text sources that are produced in several languages (e.g. translated legal documents or subtitles for the same videos in different languages). The texts are then aligned and processed to produce parallel sentences in the different languages. For instance, the CCaligned corpus contains web-document pairs in 137 languages obtained by identifying URLs that are translations of the same page \cite{el-kishky_ccaligned_2020}. However, a recent evaluation of the main available corpora has shown that in some cases up to two-third of the audited samples were misaligned \cite{kreutzer_quality_2022}. Also the reliance on sources that are produced in several languages makes it difficult to find more spontaneous texts in the dataset.

To tackle these challenges, in this paper we present the \textsc{esCorpius} Spanish corpus with the following properties:
\begin{itemize}
    \item It is cleaner than state-of-the-art corpora and deduplicated.
    \item It maintains both document and paragraph boundaries allowing language models to deal with textual data in the same way as humans do, thus unlocking the capabilities of Natural Language Generation to understand the paragraph representation.
    \item The data downloaded maintains the traceability of the origin of each document. This level of traceability makes it possible to apply the right of withdrawal of individual website owners or individual persons whose data are cited on websites and are protected by GDPR. It allows to systematically exclude blacklisted websites.
    \item In terms of size and the properties mentioned above it is the largest and cleanest Spanish corpus till the date.
\end{itemize}

\section{\textsc{esCorpius} at a glance} 

A total of 39,502 compressed WARC (Web Archive) from Common Crawl files were processed (see section \ref{sec:download} for more details). The compressed information occupied about 180 TB and the size of the processed decompressed information is estimated to be more than 0.8 PB. 
Prior to content deduplication, the downloaded corpus was composed of 106.768.594.753 words, 3.129.248.875 lines and 163.518.405 web pages. The deduplicated and cleaned corpus size is 346.262.072.705 bytes (322.5 GB), with 104.073.706 total number of lines, 50.040.055.322 tokens, 1.125.798.968 paragraphs and 2.421.598.201 sentences.

The corpus has been uploaded to HuggingFace in 33 chunk files, with a size of 10 GB each one of them, to facilitate download and data processing. The Data Availability Section contains the link to the resource.

As argued in \cite{10.1145/3442188.3445922}, corpora for language modelling should not only be compared in terms of size, but also in the quality and traceability of their data. Table \ref{table:1} shows a comparison of \textsc{esCorpius} with the main state-of-the-art datasets in Spanish or that have excerpts in Spanish (ES). This comparison has been performed not only in terms of size, but also encompassing quality factors related to language identification, parsing, leaning and deduplication that will be further explained in the following section. For multilanguage datasets, the numbers reported in the table correspond to the Spanish samples only. The Table excludes the Spanish National Library corpus \cite{PLN6405} as it is not available at the time of writing this paper.

\begin{table}[h!]
\centering
\begin{tabular}{l p{2cm} p{2cm} p{2cm} p{2cm} p{2cm}}
\hline
 & \textbf{OSCAR} 22.01 \cite{2022arXiv220106642A}  &  \textbf{mC4} \cite{xue_mt5_2021} & \textbf{CC-100} \cite{wenzek_ccnet_2020} & \textbf{ParaCrawl} v9 \cite{banon-etal-2020-paracrawl} & \textbf{esCorpius}
(ours) \\ 
\toprule
Size (ES) & \cellcolor[gray]{0.9} 381.9 GB & \cellcolor[gray]{0.9} 1,600 GB & 53.3 GB & 24.0 GB & \cellcolor[gray]{0.9} 322.5 GB\\ \hline
Docs (ES) & 51M & \cellcolor[gray]{0.9} 416M & -  & - & \cellcolor[gray]{0.9} 104M\\ \hline
Words (ES) & \cellcolor[gray]{0.9} 42,829M & \cellcolor[gray]{0.9} 433,000M & 9,374M & 4,374M & \cellcolor[gray]{0.9} 50,773M \\ \hline 
Lang. identification & fastText & CLD3 & fastText & CLD2 & \cellcolor[gray]{0.9} CLD2 + fastText \\ \hline
Elements & Document & Document & Document & Sentence & \cellcolor[gray]{0.9}Document and paragraph\\ \hline
Parsing quality & Medium & Low & Medium &\cellcolor[gray]{0.9} High & \cellcolor[gray]{0.9} High \\ \hline
Cleaning quality & Low & No cleaning & Low & \cellcolor[gray]{0.9} High & \cellcolor[gray]{0.9} High \\ \hline
Deduplication & No & No & No & \cellcolor[gray]{0.9}Bicleaner & \cellcolor[gray]{0.9} dLHF \\ \hline
Language & Multilanguage & Multilanguage & Multilanguage & Multilanguage & Spanish \\ \hline
Licence & CC BY 4.0 & OCD-BY 1.0 & Common Crawl & CC0 & CC-BY-NC-ND 4.0 \\ 
\hline
\end{tabular}
\caption{Comparison of the main state-of-the-art Spanish corpora or Spanish excerpts of multilingual corpora}
\label{table:1}
\end{table}

\section{Data download \& cleaning process}

To generate \textsc{esCorpius} we have processed WARC files from Common Crawl. We have ideated and implemented a novel cleaning pipeline that allows obtaining high quality data, which we have applied to the WARC files in order to obtain a clean Spanish dataset. 

\subsection{Common Crawl repository}
Common Crawl is a web archive that contains petabytes\footnote{See 
\url{https://commoncrawl.github.io/cc-crawl-statistics/plots/crawlsize} to explore current crawling sizes.} of data collected since 2008. The repository contains raw web page data in the WARC file format, request/response metadata files in WAT format, and text data contents extracted from WARC and stored in WET files. 

Some corpora such as mC4 use WET files to generate the corpus content. However, this has several disadvantages. The main problem with WET is that the process followed to remove HTML tags and extract the text is error-prone. After HTML tags are removed, frequently the text is either divided into unconnected pieces or merged with other unrelated textual information. For example, the sentence ``We offer <b>fast</b> transportation'' could be divided into \{ ``We offer'', ``fast'', and ``transportation''\}, thus loosing the integrity of the original text.

Additionally, the text extractions from WET files are HTML content agnostic in the sense it is not checked whether the content belongs to less relevant parts of the web (e.g. a menu or a footer).

Hence, for our work we decided to use the original WARC files as some transformations made for creating the WETs are irreversible or, at least, very costly to repair.

The reason why related work chooses WET \cite{Brown2020LanguageMA, Raffel2020ExploringTL} is that in WET format there is a close relation between the file size and the amount of textual data it contains: nearly 100\% of the contents of WET files is pure text. On the contrary, the textual data and WARC file size ratio is low, and WARC files require parsing to obtain the text, which takes longer computing times and resources.

\subsection{WARC files and archiving standard}

Common Crawl distributes contents in different folders (prefixes in Amazon S3 terminology) at a rate of one folder per month since August 2014.\footnote{Prior to 2014 several months were stored in the same folder.} For each month, a growing number of WARC files are generated, currently reaching more than 72,000\footnote{According to \url{https://commoncrawl.org/the-data/get-started/}}.

The size of the WARC files is variable: they are published compressed in Gzip compression format to save space in the repository with a size that, since January 2015, ranges from 0.9GB to 1.1GB. The compression ratios observed in the files are between 4:1 to 5:1, so the final size of these files, once decompressed, ranges from 3.9GB to 5.2GB.

WARC is an extension of the ARC file format (ARC) used to store "web crawls" as sequences of content blocks. WARC (Web ARChive) file format concatenates multiple resource records (data objects), consisting of simple text headers and data blocks into a long file. WARC format is a preservation format defined by the International Internet Preservation Consortium (IIPC). The WARC format offers a standardized way to manage billions of web collected resources.\footnote{See the following web page for more information on this standard file format: \url{https://www.iso.org/obp/ui/\#iso:std:iso:28500:ed-2:v1:en}}

\subsection{Common Crawl subcorpus selection and cleaning}
\label{sec:download}

A Common Crawl Subcorpus of WARC files from the period 2015-2022 has been selected to guarantee stability in the file format and content encoding. The compressed information occupied about 180 Tb and the size of the processed decompressed information is estimated to be more than 0.8 Pb. Each WARC file is usually divided on 100 segment files. A number of 39,502 segments of compressed WARC files were processed, so we have just computed a part of the two hundred available segments on Common Crawl.

For each one of the CPU cores employed in the generation of \textsc{esCorpius} the following protocol was performed:
\begin{enumerate}
    \item Download a WARC from Common Crawl.
    \item Open a Gzip file reader.
    \item While reading the Gzip file, partially parse the WARC format.
    \item Parse the webpage and fix the encoding to UTF-8.
    \item Obtain the language used in the document (see section \ref{sec:langid}). Proceed if the language detected is Spanish.
    \item Extract the text that is correct.
    \item Store the record in the format described in section \ref{sec:outformat}.
\end{enumerate}

In order to avoid obtaining too much duplicated content and/or content which is very similar (e.g. COVID-19 content) we randomized the WARC URL order that is fed to the cleaning cluster. Also we performed a deduplication process over the data obtained that is described in section \ref{sec:deduplication}.

\subsection{Language detection process}
\label{sec:langid}

After extracting the text from Common Crawl, it was neccessary to select only text in Spanish. Language detection is a very important part of the pipeline to produce accurate results. However, many times the language identificaiton methods used are not robust and text in languages different from the target one are considered in the datasets. In the case of Spanish, it can be easily confused with other romance languages.

To avoid such mistakes and produce more robust results, we have carried language identification in two steps. First, a quick filtering based on the Compact Language Detector 2 (CLD2) tool\footnote{See \url{https://github.com/CLD2Owners/cld2} for more details.} has been used. Secondly, we used the fastText tool \footnote{\url{https://github.com/facebookresearch/fastText}}\cite{bojanowski2017enriching} which requires larger computational resources in order to verify the language identification made by CLD2. In the creation of the corpus, the criterion of quality prevailed over that of quantity. We have sacrificed corpus length and processing time in exchange for greater certainty that the language of the text is actually Spanish.

\subsection{Output storage format}
\label{sec:outformat}
The output format of this process and the format in which the corpus is distributed is JSONL. For each line there is a separate JSON document with the following fields:
\begin{itemize}
    \item \texttt{id}: Unique document identifiers UUIDv4\footnote{The complete specification of UUIDv4 can be found in \url{https://www.ietf.org/rfc/rfc4122.txt}} over the whole corpus.
    \item \texttt{text}: textual content of the paragraph.
    \item \texttt{url\_warc}: this is the identifier of the WARC file from which the web page from which the text has been extracted following the CommonCrawl segments nomenclature ("\texttt{s3://commoncrawl/crawl-data/
    CC-MAIN-<YYYY>-<MM>/segments/<id>/warc/CC-MAIN-<id>.warc.gz}", where YYYY is the 4 digits year and MM the WARC archive month).
    \item \texttt{url}: URL address from which the text has been extracted.
\end{itemize}
A sample corpus JSON register is shown below. Note that currently the article referenced in the example does not exist although it is archived on Common Crawl.

\begin{lstlisting}
{
    "id":"8280bafd-5984-4a5e-8436-af56a474d9cd", 
    "text":"<textual content>",
    "url_warc":"s3://commoncrawl/crawl-data/CC-MAIN-2019-04/segments/
        1547583730728.68/warc/CC-MAIN-20190120184253-20190120210253-00091.warc.gz",
    "url":"http://alertatierra.com/continente/61-noticias/volcanes/
        2135-erupcion-de-turrialba-y-rincon-de-la-vieja-en-costa-rica"
}
\end{lstlisting}

As described in \cite{lee2021deduplicating}, it is crucial to generate datasets for training language models that are free of duplicity. This makes it possible to train models that do not memorise sentences due to their high degree of duplicity, which artificially results in fewer training steps and higher accuracy. Using deduplication, the overlap between training, validation and test sets is reduced, improving the training procedure as well as the confidence on the model proficiency.

We do not perform any content filtering process based on URLs as we want to avoid any kind of censorship. We expect users to perform this filtering if they are interested on it.\footnote{We encourage users to understand the license terms before doing so.} Additionally, depending on the purpose of the model, the corpus can be filtered out to show only the results from specific URLs. 

Our novel deduplicacion process, which we have named dLHF, is comprised of two main steps. First of all, we deduplicate complete contents of the corpus via exact matching at document level. Subsequently, we perform the same deduplication at paragraph level. In order to deal with paragraphs, the text is normalized and noisy information is removed. Even though the process may appear naïve, the greatest part of the deduplication happens in this step.

After that, we perform a more complex deduplication based on Local Sensitive Hashing. We adapted state-of-the-art code to work in parallel, use less computational resources and, more importantly, to avoid unstructuring the document. This last feature is highly relevant as we have found that some of the deduplication software available performs operations that break the document integrity. Actually, the fact that the OSCAR corpus suffers this issue, leads us to think that it was not subject to deduplication whatsoever.\footnote{See the last point of the Section "Changes" of the following URL for more details: \url{https://oscar-corpus.com/post/oscar-v22-01/}}

\section{Technical environment used for corpus generation}

The selected massively parallel processing infrastructure relies on open source solutions for data analytics, Apache Hadoop and Apache Spark framework, integrated in the Amazon Elastic Map Reduce (EMR) service of Amazon Web Services (AWS). In particular, the following application stack has been used:

\begin{itemize}
\item \texttt{Hadoop File System (HDFS)}, a distributed and scalable file system, was used as an auxiliary repository in the tasks executed in EMR, which is built with part of the cluster nodes (Core Nodes).

\item \texttt{Amazon Simple Storage Service (S3)}, in particular EMR FS, which is a Hadoop reimplementation of AWS object storage.

\item \texttt{Apache Spark}, this open source engine was used for the parallelisation of data cleaning tasks. The parallel WARCs processor is based on PySpark.

\item \texttt{Apache YARN} was used as resource manager used for scheduling and monitoring the execution of tasks in the EMR cluster.

\item \texttt{Ganglia} has been used for monitoring the cluster status. Cluster monitoring is important in the early stages, for node load visualization and selecting the number and type of nodes used in the cluster.
\end{itemize}

For the EMR cluster we used three types of nodes: 1 Master Node, which is the orchestrator of the cluster and on which YARN resource manager runs; 3 Core Nodes, which support the local HDFS file system; and large collection of Task Nodes, which are nodes dedicated exclusively to the execution of tasks. All of these nodes are supported by virtual compute instances (similar to the concept of a virtual machine in a AWS data centre environment) thanks to the Amazon Elastic Compute Cloud (EC2) service.

Amazon EMR supports different types of EC2 instances depending on the characteristics of the task to be executed. For the corpus generation process, general purpose instances have been considered, although instances with GPU and advanced networking capabilities are also available. In particular, we used the following EC2 instance types:

\begin{itemize}
\item \texttt{M5 instances}\footnote{\url{https://aws.amazon.com/ec2/instance-types/m5/}}: general purpose instances powered by Intel Xeon Platinum 8000 series processors up to 3.1 GHz. They have a network bandwidth ranging from 10 to 25 Gbps depending on the selected size.
\item \texttt{R5 instances}\footnote{\url{https://aws.amazon.com/ec2/instance-types/r5/}}: memory-optimised instances, with the same type of processors as the M5 instances but with a vCPU:RAM ratio of 1:8, allowing more memory-demanding tasks to be executed.
\end{itemize}

\subsection{Textual content deduplication infrastructure}
Due to the structure of the problem and the large memory requirements, the paradigm of the Hadoop cluster was not a good fit for the deduplication process.

For our deduplication tools we used a machine with Intel Xeon Platinum 8176 2.10GHz processor (112 processor threads) and 1.5TB of RAM memory. With such amount of RAM memory we could deduplicate the corpus efficiently in less than 3 hours.

\section{Conclusions and future work}
This paper presents \textsc{esCorpius}, a massive cleaned Spanish web crawl corpus, which has been produced by means of a novel and effective approach to produce high quality deduplicated corpora from Common Crawl. The collected data fully respects the document and paragraph boundaries in order for language models to be more accurate. The document source URLs allow full traceability of the data which permits creating domain-specific Language Models, indexing data, complying with EU regulation (such as the right to be forgotten). 

As future work, we suggest increasing the time to extract the corpus as, according to our calculations, there is the chance to still extract a corpus 200 times bigger. As this corpus is only a crawling based corpus, we propose to extend it and create a The Pile \cite{DBLP:journals/corr/abs-2101-00027} like Spanish corpus. We have shared the corpus in HuggingFace with the hope that its potential users can contribute to the advancement of NLP technologies in Spanish by performing different analyses of the corpus (e.g. topic modelling) and creating new language models and embeddings for the community.

\section*{Acknowledgements}
We want to thank Amazon Web Services Spain, specially, to José (Pepe) López Rodríguez and Alberto González Dueñas for their help on setting up the cluster and managing the communication with the Common Crawl AWS team. This work could not have been possible without their invaluable help.

\section*{Data availability}
LHF Labs has released \textsc{esCorpius} under CC-BY-NC-ND 4.0, \footnote{\url{https://creativecommons.org/licenses/by-nc-nd/4.0/}}, license in HuggingFace: \url{https://huggingface.co/datasets/LHF/escorpius}. This license allows reusers to copy and distribute the material in any medium or format in unadapted form only, for noncommercial purposes only, and only so long as attribution is given to the creator.

\bibliographystyle{unsrt}  
\bibliography{references}

\end{document}